\documentclass{article}

\usepackage[nonatbib,preprint]{camera_ready} 

\usepackage[utf8]{inputenc} 
\usepackage[T1]{fontenc}    
\usepackage{hyperref}       
\usepackage{url}            
\usepackage{booktabs}       
\usepackage{amsfonts}       
\usepackage{nicefrac}       
\usepackage{microtype}      
\usepackage{xcolor}         

\usepackage{graphicx}
\usepackage{subfigure}

\usepackage{amsmath}
\usepackage{amssymb}
\usepackage{wrapfig}
\usepackage{placeins}

\usepackage[sorting=none,style=ieee,citestyle=numeric-comp, backend=biber]{biblatex} 
\def\MI[#1]#2{I_{#1}(#2)}

\newcommand{\xPlus}{\boldsymbol{x}_{m:n}}
\newcommand{\xMinus}{\boldsymbol{x}_{1:m-1}}
\newcommand{\fForward}{F^{-1}_{\theta(\xMinus, \xPlus)}(\boldsymbol{x}_{m:n})}
\newcommand{\fInverse}{F_{\theta(\xMinus, \xPlus)}(\boldsymbol{z})}

\newcommand*\dif{\mathop{}\!\mathrm{d}}

\title{Funnels\\ \large{Exact maximum likelihood with dimensionality reduction}} 

\author{%
  Samuel Klein \\
  University of Geneva\\
  \texttt{samuel.klein@unige.ch} \\
  \And
  John A. Raine \\
  University of Geneva\\
  \texttt{john.raine@unige.ch} \\
  \AND
  Sebastian Pina-Otey \\
  University of Geneva\\
  \texttt{sebastian.pinaotey@unige.ch} \\
  \And
  Slava Voloshynovskiy \\
  University of Geneva\\
  \texttt{svolos@unige.ch} \\
  \And
  Tobias Golling \\
  University of Geneva\\
  \texttt{tobias.golling@unige.ch} \\
}

\begin{document}

\maketitle

\begin{abstract}

Normalizing flows are diffeomorphic, typically dimension-preserving, models trained using the likelihood of the model. We use the SurVAE framework to construct dimension reducing surjective flows via a new layer, known as the funnel. We demonstrate its efficacy on a variety of datasets, and show it improves upon or matches the performance of existing flows while having a reduced latent space size. The funnel layer can be constructed from a wide range of transformations including restricted convolution and feed forward layers.
\end{abstract}

\section{Introduction}

Normalizing flows~\cite{kingma2018glow, tabak2013family, real_nvp, durkan2019neural, papamakarios2017masked, papamakarios2021normalizing} are powerful generative models that can be used to approximate the likelihood of data directly, unlike variational methods which calculate a lower bound~\cite{kingma2014autoencoding, rezende2014stochastic}. Flows are restricted to being diffeomorphisms and, unless the data is known to lie on a Riemannian manifold \cite{papamakarios2021normalizing}, preserve the dimensionality of the data. This leads to undesirable effects such as out of distribution samples being assigned higher likelihoods than inliers \cite{nalisnick2019deep} and difficulties in modelling complex high dimensional data \cite{krause2021caloflow,kingma2018glow}. These issues stem, at least in part, from the curse of dimensionality. Furthermore, autoregressive flows are the most flexible family of flows, and the cost of inverting these models scales linearly with the dimension of the data~\cite{papamakarios2021normalizing}. Therefore, methods for generalizing flows to be dimensionality reducing are required.

In contrast to strictly invertible functions, like those used in flows, surjective transformations can be used to exactly calculate the likelihood while changing the dimension of the data. The SurVAE framework can be used to build modular surjective transformations that allow for exact likelihood inference~\cite{nielsen2020survae}. This framework provides a bridge between variational autoencoders (VAEs), which can perform arbitrary dimension reduction but only provide a lower bound for the likelihood, and flows which are typically dimension preserving diffeomorphisms. Within the SurVAE framework the log likelihood of data $\boldsymbol{x} \in \mathcal{X}$ is modelled as
\begin{equation*}
    \log (p(\boldsymbol{x} )) = \log (p_\theta(\boldsymbol{z})) + \mathcal{V}(\boldsymbol{x}, \boldsymbol{z}) + \mathcal{E}(\boldsymbol{x}, \boldsymbol{z}), ~~\boldsymbol{z} \sim q (\boldsymbol{z} | \boldsymbol{x}), ~~\boldsymbol{z} \in \mathcal{Z},
\end{equation*}
where $q (\boldsymbol{z} | \boldsymbol{x})$ is an amortized variational distribution and $p_\theta$ is the base distribution defined on the latent space $\mathcal{Z}$. The $\mathcal{V}$ term is referred to as the likelihood contribution and $\mathcal{E}$ as the bound looseness.

The likelihood contribution for a deterministic surjection $f: \mathcal{X}\rightarrow\mathcal{Z}$ is given by
\begin{equation}
    \mathcal{V}(\boldsymbol{x}, \boldsymbol{z}) = \lim_{q(\boldsymbol{z}|\boldsymbol{x}) \rightarrow \delta (\boldsymbol{z} - f(\boldsymbol{x}))} \mathbb{E}_{q(\boldsymbol{z}|\boldsymbol{x})} \left[ \log \frac{ p(\boldsymbol{x} | \boldsymbol{z}) }{ q(\boldsymbol{z}|\boldsymbol{x}) } \right],
    \label{eq:likelihood_contribution}
\end{equation}
where $p(\boldsymbol{x} | \boldsymbol{z})$ is a stochastic generative process. The bound looseness term is zero if the function $f(\boldsymbol{x})$ satisfies the \textit{right inverse condition} (\ref{app:defs_and_notation}).

The SurVAE paper~\cite{nielsen2020survae} provides useful novel layers but no principled method for directly extending flows. In this work we derive a layer which extends flows to (i) be surjective with exact likelihoods; (ii) have latent spaces with smaller dimensions than the data space and (iii) propagate compressed information explicitly. Our method also allows new types of invertible transformations to be derived, which we demonstrate by converting convolutional neural networks and multilayer perceptrons into efficient exact maximum likelihood estimators (\ref{app:funneling_standard_layers}). Deriving such transformations allows flows to be extended to large dimensional datasets. We benchmark the proposed method on a wide variety of real world datasets and demonstrate their applicability and usefulness for downstream tasks. The code required to repeat the experiments is provided at \url{https://github.com/sambklein/funnels_repo}.

\section{Exact maximum likelihood with dimension reduction}

Given $\boldsymbol{x}\in\mathcal{X}$, any diffeomorphism $F^{-1}$ that acts on a subset of the features in $\boldsymbol{x}$ as
\begin{equation}
    \boldsymbol{z} = F^{-1} \left( \xPlus \right),
\end{equation}
for $m < n = \dim({\mathcal{X}})$, satisfies the \textit{right inverse condition}. By taking $F$ to be parameterized by a function $\theta(\xMinus, \xPlus)$ a flexible new family of models can be derived. This transformation is a generalization of the tensor slicing layer~\cite{nielsen2020survae, huang2020augmented, chen2020vflow, real_nvp}, where one of the key differences is $F_\theta$ is any function that can be inverted for $\xPlus$ given $\xMinus$ (\ref{app:relation_to_slice}). It also explicitly allows the dependence of the model on $\boldsymbol{x}_{1:m - 1}$ to be controlled and propagated efficiently.

The generative transformation within the SurVAE framework is given by
\begin{equation}
    G(\boldsymbol{z}) = 
    \begin{bmatrix}
    \xMinus \sim p(\xMinus | \boldsymbol{z}) \\
    F_{\theta(\xMinus, \boldsymbol{z})} \left(  \boldsymbol{z} | \xMinus \right)
    \end{bmatrix}.
\end{equation}
The process $p(\xMinus | \boldsymbol{z})$ can be any continuous density that maps to the full range of $\xMinus$, this can be parameterized by a density $p_\phi(\xMinus | \boldsymbol{z})$. 
To derive the likelihood contribution for a funnel layer we start by following the SurVAE framework~\cite{nielsen2020survae} and using the composition of a smooth function $g$ with a Dirac $\delta$-function and a diffeomorphism $f$ where
\begin{equation}
    \int \dif \boldsymbol{x} \delta (g (\boldsymbol{x})) f (g (\boldsymbol{x})) \left| \det \frac{\partial g(\boldsymbol{x})}{\partial \boldsymbol{x}} \right|
    =
    \int \dif \boldsymbol{u} \delta(\boldsymbol{u}) f(\boldsymbol{u}),
\end{equation}
to conclude that
\begin{equation}
    \delta (g(\boldsymbol{x})) = \left| \det \frac{\partial g(\boldsymbol{x})}{\partial \boldsymbol{x}} \right|^{-1}_{\boldsymbol{x}=\boldsymbol{x}_0} \delta(\boldsymbol{x} - \boldsymbol{x}_0),
\end{equation}
where $\boldsymbol{x}_0$ is the root of $g(\boldsymbol{x})$. This assumes the root is unique, the Jacobian is non-singular and $f$ has compact support. Using this result we take $g(\boldsymbol{x}) = \boldsymbol{z} - \fForward$ where the unique root is given by $\boldsymbol{x}_0 = \fInverse$ and
\begin{align*}
    q (\boldsymbol{z} | \boldsymbol{x})  &= \delta (\boldsymbol{z} - \fForward)\\
    &= \delta (\boldsymbol{x}_{m:n} - \fInverse) \left| \mathrm{det} (J_\theta(\boldsymbol{x}_{m:n})) \right|^{-1},
\end{align*}
where
\begin{align*}
    J_\theta = \left. \frac{\partial \fForward}{\partial \xPlus} \right|_{\xPlus=\fInverse}.
\end{align*}

With a change of variables to $\boldsymbol{x}_{m:n}'=\fInverse$ such that
\begin{align*}
    \dif \xPlus' &= \dif \boldsymbol{z} \left| \det (J_\theta^{-1} (\xPlus)) \right|\\
    &= \dif \boldsymbol{z} \left| \det (J_\theta (\xPlus)) \right|^{-1},
\end{align*}
 Eq.~\eqref{eq:likelihood_contribution} can be written as
\begin{align}
    \mathcal{V}(\boldsymbol{x}) 
    &= \int \dif \boldsymbol{z} \delta \left(\boldsymbol{z} - \fForward\right) \log \left( \frac{ p_\phi(\xMinus | \boldsymbol{z}) }{ \delta \left(\boldsymbol{z} - \fForward \right) } \right) \nonumber \\
    &= \int \dif \xPlus' \delta (\xPlus - \xPlus') \log \left(  p_\phi(\xMinus | \boldsymbol{z}) \left| \det (J_\theta (\xPlus')) \right| \right) \nonumber\\
    &= \log \left(  p_\phi \left(\xMinus \big| \fForward \right) \right) +  \log\left( \left| \det (J_\theta (\xPlus)) \right| \right). 
    \label{eq:new_loss}
\end{align}

This expression can be understood as the determinant of the Jacobian from the subset of the coordinates for which the model is a diffeomorphism ($\xPlus$), plus the log likelihood of the coordinates which condition the diffeomorphism ($\xMinus$). The log likelihood acts as a reconstruction term that penalises moving away from invertibility. This is the derivation of the likelihood contribution for a funnel layer with an inference surjection; the same for a generative surjection can also be derived as shown in \ref{app:derivation_of_loss}.
%
\begin{figure}
  \begin{center}
    \includegraphics[width=0.6\textwidth]{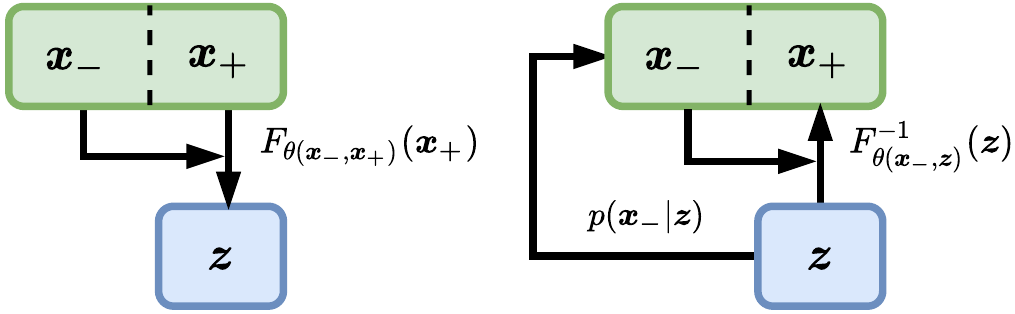}
  \end{center}
  \caption{Funnel layer with the forward (left) and inverse (right) passes.
  In each pass $\boldsymbol{x_-} = \xMinus$ conditions the diffeomorphism $F$. In the inverse pass $\boldsymbol{x_-}$ is first sampled from $p(\boldsymbol{x_-} | \boldsymbol{z})$ and then used to condition the inverse of the diffeomorphism to return a sample of $\boldsymbol{x_+} = \xPlus$.}
  \label{fig:model_sketch}
\end{figure}
The likelihood contribution in Eq.~\eqref{eq:new_loss} can be used to form models that reduce dimensions by arbitrary amounts, can be trained on exact maximum likelihood estimation and permit exact latent variable inference. This likelihood contribution defines a \textit{funnel} layer. The transformation $F_\theta$ used in this layer needs to be invertible for $\xPlus$ given $\{\xMinus, \boldsymbol{z}\}$ and have a tractable Jacobian determinant. Such transformations are abundantly available in the form of (conditional) invertible neural networks (INNs), but also appear in linear transformations such as convolutional kernels. A sketch of the layer can be seen in Fig.~\ref{fig:model_sketch}. The likelihood contribution in Eq.~\eqref{eq:new_loss} can be used to define funnel layers.

\section{Novel funnel layers}

In this section we derive some novel funnel layers to demonstrate the use of Eq.~\eqref{eq:new_loss} for deriving layers from existing methods. 

\textbf{Convolutional kernels}. Convolutional kernels on one dimensional data can be turned into funnels as follows. Given a one dimensional vector $\boldsymbol{x} \in \mathbb{R}^{n}$ the forward transformation of the funnel can be factorised to form a $2\times 1$ convolutional kernel ($[a, b]$) with stride two
\begin{equation*}
    q (\boldsymbol{z} | \boldsymbol{x}) = \prod_{i=1}^{n/2} q (z_i | x_{2i - 1}, x_{2i})
\end{equation*}
with
\begin{align*}
   z_i  &= F^{-1}_{\theta(x_{2i - 1}, x_{2i})}(x_{2i - 1}) \\
    &= a x_{2i - 1} + b x_{2i}
\end{align*}
where $[a, b]$ are the parameters of the convolutional kernel. Replacing $\xMinus$ with $x_{2i - 1}$ in Eq.~\eqref{eq:new_loss} the Jacobian of each kernel is given by
\begin{equation*}
    J_i = a,
\end{equation*}
and the total Jacobian is given by $\frac{n}{2}a$. The total likelihood contribution in this case is given by
\begin{align}
    \mathcal{V}(\boldsymbol{x}) = \log \left(  p_\phi \left( \{ x_{2i} \}_{i=1}^{n/2} \big| \boldsymbol{z} \right) \right) + \log\left(\frac{n}{2}| a |\right).
    \label{eq:conv_contr}
\end{align}
Given $x_{2i}' \sim p_\phi \left( \{ x_{2i} \}_{i=1}^{n/2} \middle| \boldsymbol{z} \right)$, the kernel $[a, b]$, and $\boldsymbol{z}$ we can find
\begin{equation}
    x_{2i - 1} = J_i^{-1} (z_i - b x_{2i}'),
\end{equation}
for all $i$ such that the right inverse condition is satisfied. Here the different elements of the kernel play different roles in the layer, with $a$ appearing directly in the loss and $b$ implicitly in $p_\phi$. Choosing $x_{2i - 1}$ to replace $\xPlus$ was arbitrary. These considerations mean some features in the input vector are treated differently, partially breaking the spatial invariance of convolutional kernels. Details of how this is generalised can be found in \ref{app:conv_kernels} 

\textbf{MLP funnels}. Non-invertible activation functions such as ReLU~\cite{hahnloser2000digital} can be used with exact maximum likelihood techniques \cite{nielsen2020survae}. Therefore the main challenge of forming a funnel multilayer perceptron (MLP) lies in parameterising the weight matrix of an MLP such that it can be used with exact maximum likelihood estimation.

The weight matrix $\boldsymbol{W} : \mathcal{X}^{n} \rightarrow \mathcal{Z}^{n - m}$ of a dimension reducing MLP ($0 < m < n$) can be parameterized by an invertible block matrix $\boldsymbol{R}_{m, m}$ and a generic matrix $\boldsymbol{W}'_{m, n - m}$ such that
\begin{equation*}
    \boldsymbol{W} \boldsymbol{x} = \boldsymbol{R} \xPlus + \boldsymbol{W}' \xMinus,
\end{equation*}
and the likelihood contribution of the transformation becomes
\begin{equation*}
    \mathcal{V}(\boldsymbol{x}) = \log \left(  p_\phi \left(\xMinus \big| \boldsymbol{W} \boldsymbol{x} + \boldsymbol{b} \right) \right) +  \log\left( \left| \det (\boldsymbol{R}) \right| \right),
\end{equation*}
where $\boldsymbol{b}$ is the bias of the layer. Here the matrix $\boldsymbol{R}$ can have a relatively complex determinant as it only needs to be calculated once per batch update. Also it does not need to be easily invertible, as a lower bound can be found for the likelihood by violating the right inverse condition and instead optimising
\begin{equation*}
    \mathcal{V}(\boldsymbol{x}) = \log \left(  p_\phi \left(\boldsymbol{x} \big| \boldsymbol{W} \boldsymbol{x} + \boldsymbol{b} \right) \right) +  \log\left( \left| \det (\boldsymbol{R}) \right| \right).
\end{equation*} 

This transformation also treats some coordinates differently by imposing that the submatrix $\boldsymbol{R}$ is invertible. This can be partially alleviated by placing flow layers or a fixed random permutation before applying the weight matrix to ensure no specific coordinates are selected to be treated differently. Using results from the existing literature the matrix $\boldsymbol{R}$ can be parameterized with the LU decomposition \cite{oliva2018transformation} or the QR decompostion~\cite{hoogeboom2019emerging}. Dimension increasing and preserving MLPs are discussed in \ref{app:MLPfunnels}.

\section{Experiments}

%
\begin{figure} 
  \begin{center}
    \includegraphics[width=0.65\textwidth]{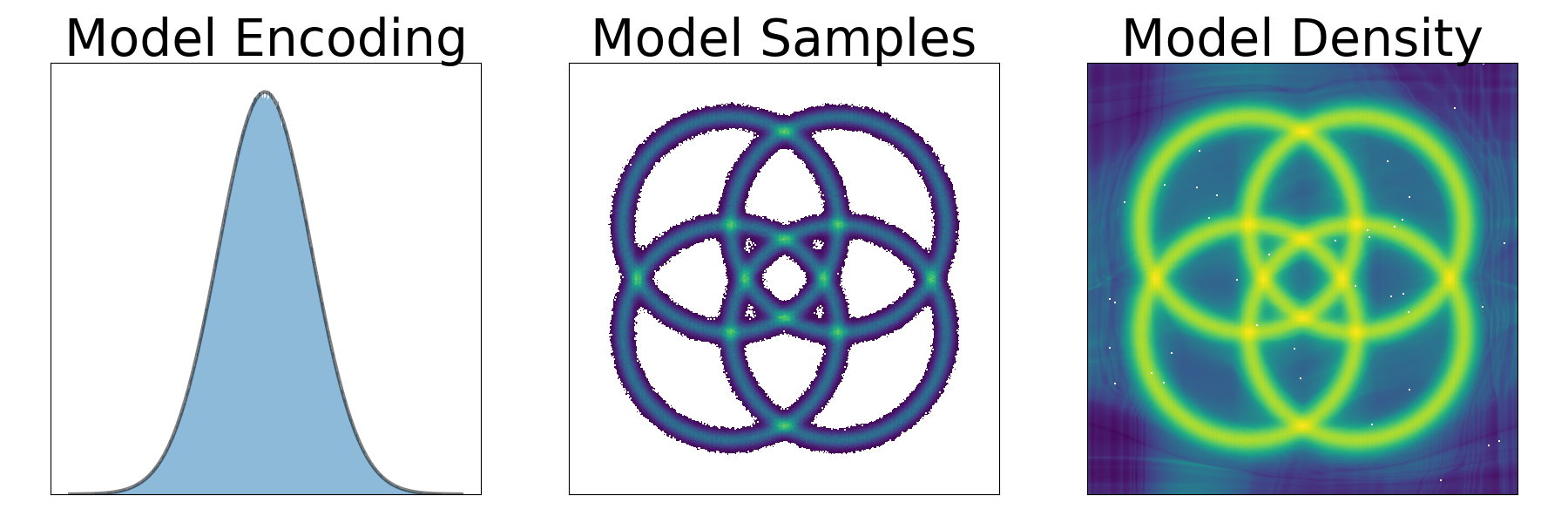}
  \end{center}
  \caption{A funnel NSF. The black line in the left figure is a true standard normal distribution and in blue is a histogram of the model encoding. Samples drawn from the prior (middle) and model likelihood (right).}
  \label{fig:quantitative}
\end{figure}
\textbf{Synthetic Data.} We demonstrate the method works qualitatively in Fig.~\ref{fig:quantitative} where a funnel layer is used to extend a neural spline flow (\textsc{NSF})~\cite{durkan2019neural} to map from a complex two dimensional distribution to a simple one dimensional distribution (\ref{app:plane_data}). The matching of the encoded data to the base distribution can be seen on the left, and a faithful distribution of model samples in the middle and the likelihood on the right. The success of the funnel model on this highly correlated dataset demonstrates its efficacy in handling complex data. 

\textbf{Tabular Data.} We next evaluate the performance on selected UCI datasets~\cite{uci_datasets} and the BSDS$300$ collection of natural images~\cite{MartinFTM01} with the same preprocessing and experimental setup as used in masked autoregressive flows~\cite{papamakarios2017masked}, as well as utilising the same data and preprocessing~\cite{papamakarios_george_2018_1161203}. The extensions made to the masked autoregressive flows in Ref.~\cite{durkan2019neural} are applied to the models here. 
We replicate the experimental setup of NSF models~\cite{durkan2019neural} as baselines, and extend all models to become funnels as outlined below. No comparisons to SurVAE layers are considered as there are no models that offer comparable performance except for the slice surjection, which we consider to be a simplified instance of a funnel layer where $F$ is the identity (\ref{app:relation_to_slice}). A baseline VAE model constructed of multilayer perceptrons (MLPs) is also included where the decoder outputs both a mean and standard deviation such that the evidence lower bound (ELBO)~\cite{kingma2014autoencoding} can be used as a bound for the likelihood (\ref{app:vae_elbo}).

\begin{table*}[h]
    \centering
    \caption{Test log likelihood (in nats, higher is better) for UCI datasets and BSDS300, with error bars corresponding to two standard deviations. All models are coupling flows. The \textsc{F} prefix signifies the model has been converted into a funnel with a $25\%$ smaller latent space.
    }
    \resizebox{\textwidth}{!}{
    \begin{tabular}{lccccc}
        \toprule \textsc{Model} & \textsc{POWER} & \textsc{GAS} & \textsc{HEPMASS} & \textsc{MINBOONE} & \textsc{BSDS$300$}\\
        \cmidrule(r){1-6} \textsc{Glow} & $\boldsymbol{0.38 \pm 0.01}$ & $12.02 \pm 0.02$ & $\boldsymbol{-17.22 \pm 0.02}$ & $-10.65 \pm 0.45$ & $\boldsymbol{156.96 \pm 0.28}$ \\
        \textsc{F-Glow} & $\boldsymbol{0.38 \pm 0.01}$ & $\boldsymbol{12.03 \pm 0.02}$ & $-18.25 \pm 0.02$ & $\boldsymbol{-10.58 \pm 0.44}$ & $155.85 \pm 0.85 $ \\ 
        \cmidrule(r){1-6} \textsc{NSF} & $\boldsymbol{0.63 \pm 0.01}$ & $\boldsymbol{13.02 \pm 0.02}$ & $\boldsymbol{-14.92 \pm 0.02}$ & $-9.58 \pm 0.48$ & $\boldsymbol{157.61 \pm 0.28}$ \\
        \textsc{F-NSF} & $0.62 \pm 0.02$ & $12.88 \pm 0.02$ & $-15.07 \pm 0.02$ & $\boldsymbol{-9.50 \pm 0.48}$ & $156.90 \pm 0.28$ \\
        \cmidrule(r){1-6} \textsc{VAE} & ${0.10 \pm 0.02}$ & ${0.02 \pm 0.05}$ & ${-22.00 \pm 0.02}$ & $-15.39 \pm 0.49$ & ${143.70 \pm 0.29}$ \\
        \textsc{F-MLP} & $\boldsymbol{0.36 \pm 0.01}$ & $\boldsymbol{7.94 \pm 0.03}$ & $\boldsymbol{-17.80 \pm  0.02}$ & $\boldsymbol{-12.81 \pm 0.55}$ & $\boldsymbol{153.26 \pm 0.28}$ \\ 
        \bottomrule
    \end{tabular}
    }
    \label{tab:tablular_data_results}
\end{table*}


The results of the different models are shown in Table~\ref{tab:tablular_data_results}. In all tables bold numbers indicate the best model on each measure. The funnel models demonstrate competitive performance on all datasets while having a latent space that is $75\%$ the size of the input data dimension.
The dimensionality is reduced by replacing one flow layer with a funnel (\ref{app:uci_details}).
For the BSDS$300$ dataset a diagonal Gaussian ansatz is used to parameterize $p(\xMinus | \boldsymbol{z})$ and for all other datasets a flow is used.
The former ansatz results in a slight speed up in the amount of time per training step, and the latter in an increase of up to 20\% in time per training step over the baseline equivalents. Notably the VAE likelihood estimate provided by the ELBO shows significantly degraded performance for the same compression as the funnels. 

A simple funnel MLP (\textsc{F-MLP}) with a spline activation function performs much better at the density estimation tasks in Table.~\ref{tab:tablular_data_results} than a VAE while taking a similar amount of time to train per training step, which is up to six times faster than the flows considered here. These layers can be dramatically improved by combining them with simple flows and selecting the model architecture more carefully, without sacrificing training speed. An \textsc{F-MLP} layer offers a way to build fast performant models for density estimation tasks.

These results show that funnels can be used to model complex datasets from different domains while reducing the size of the latent space with respect to standard flows. The reduced latent space is useful for down stream tasks, as demonstrated in the next section on image generation and anomaly detection tasks.


\textbf{Image Data.} We benchmark funnel models on the CIFAR-$10$ \cite{Krizhevsky09learningmultiple} and downsampled ImageNet$64$ \cite{russakovsky2015imagenet, ziegler2019latent} image datasets with original $8$-bit and reduced $5$-bit colour depth.  In Table~\ref{tab:image_data_results} we demonstrate that funnel models with a similar number of parameters have similar performance to the benchmark \textsc{NSF} \cite{durkan2019neural} and baseline flow with affine transformations, while having a latent space dimension one sixteenth of the size. We also consider a VAE with a ResNet$18$ \cite{he2015deep} style encoder and decoder with a learnable standard deviation for each channel at the output of the decoder.

\begin{figure}[h!]
    \centering
    \resizebox{\textwidth}{!}{
     \begin{tabular}{cc}
        \includegraphics[scale=1]{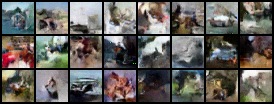} & \includegraphics[scale=1]{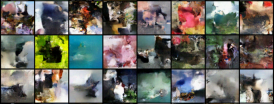} \\
    \end{tabular}
    }
    \caption{Samples drawn from a convolutional funnel model for CIFAR-$10$ $5$-bit left and ImageNet$64$ $5$-bit right.}
    \label{fig:image_samples}
\end{figure}

There are many ways to build funnel layers for images; here we use standard convolutional kernels (\ref{app:image_exp_dets}) where the kernel stride is equivalent to the width, which together with the MaxPool layer \cite{nielsen2020survae} can be used to turn convolutional neural networks into funnels. The funnel layers here have a diagonal Gaussian ansatz for $p(\xMinus | \boldsymbol{z})$, which is parameterized by a neural network. Each tile of the convolution is reconstructed separately, with an amortized ansatz for all tiles. Samples drawn from funnel models are shown in Fig.~\ref{fig:image_samples}.
%
%
%
\begingroup
\setlength{\tabcolsep}{6pt} 
\renewcommand{\arraystretch}{1}
\begin{table*}[h!]
    \centering
    \caption{Test-set bits per dimension (BPD) and latent space size (\textsc{LS}) for CIFAR-$10$ and ImageNet$64$ models, lower is better for both measures. The \textsc{F} prefix signifies the model has been converted into a convolutional funnel model. *Results taken from van den Oord et al. \cite{oord2016pixel} and Kingma et. al \cite{kingma2018glow} respectively.}
    \resizebox{\textwidth}{!}{
    \begin{tabular}{lcccccccc}
        \toprule & \multicolumn{2}{c}{\textsc{CIFAR-$10$ $5$-bit}} & \multicolumn{2}{c}{\textsc{CIFAR-$10$ $8$-bit}} & \multicolumn{2}{c}{\textsc{ImageNet$64$ $5$-bit}} & \multicolumn{2}{c}{\textsc{ImageNet$64$ $8$-bit}} \\
        \cmidrule(r){2-3} \cmidrule(r){4-5} \cmidrule(r){6-7} \cmidrule(r){8-9} \textsc{Model} & \textsc{BPD} & \textsc{LS} & \textsc{BPD} & \textsc{LS} & \textsc{BPD} & \textsc{LS} & \textsc{BPD} & \textsc{LS} \\
        \cmidrule(r){1-1} \cmidrule(r){2-9} 
        \textsc{Baseline} & $\bf{1.70}$ & $3,072$ & ${3.41}$ & $3,072$ & $1.81$ & $12,288$ & $3.91$ & $12,288$ \\
        \textsc{NSF} & $\bf{1.70}$ & $3,072$ & $\bf{3.38}$ & $3,072$ & $\bf{1.77}$ & $12,288$ & $\bf{3.82}$ & $12,288$ \\
        \textsc{F-NSF} & $1.71$ & $\bf{192}$ & $3.46$ & $\bf{192}$ & ${1.80}$ & $\bf{768}$ & $3.90$ & $\bf{768}$ \\
        \textsc{VAE} & $4.09$ & $\bf{192}$ & $7.45$ & $\bf{192}$ & $3.72$ & $\bf{768}$ & $6.67$ & $\bf{768}$ \\
        \cmidrule(r){2-9} \textsc{Glow}* & $1.67$ & $3,072$ & $3.35$ & $3,072$ & $1.76$ & $12,288$ & $3.81$ & $12,288$ \\
        \bottomrule
    \end{tabular}
    }
    \label{tab:image_data_results}
\end{table*}
\endgroup

\begin{figure}[h]
    \centering
    \resizebox{\textwidth}{!}{
     \begin{tabular}{c}
        \includegraphics[scale=1]{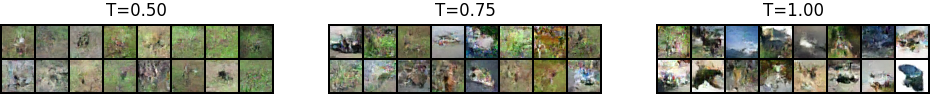} \\
        \includegraphics[scale=1]{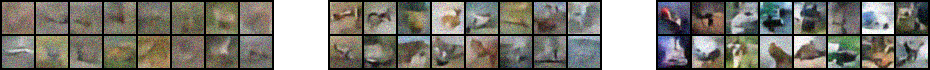} \\
        \includegraphics[scale=1]{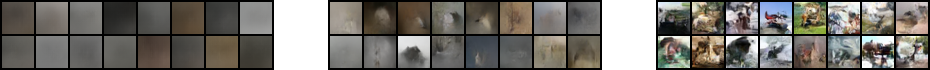} \\
    \end{tabular}
    }
    \caption{Samples drawn from a funnel model with a latent space of $48$ dimensions and BPD of $3.7$ (top), a VAE with latent dimension $48$ (middle), and an NSF model (bottom) for CIFAR-$10$ $5$-bit.}
    \label{fig:additional_image_samples}
\end{figure}

The base distribution ($p_\theta$ in Eq.~\eqref{eq:likelihood_contribution}) is chosen to be a standard normal distribution, and we remark that in all cases samples drawn from closer to the centre of the distribution have more structure in funnel models than their baseline equivalents, suggesting the encoded distribution more closely matches the base distribution as expected given the shape of the typical set of high dimensional normal distributions. In Fig.~\ref{fig:additional_image_samples} we show samples from a trained NSF model and a funnel model where samples are generated from a normal distribution at different `temperatures' (T) where samples are generated by sampling from a modified prior $\boldsymbol{z} \sim \mathrm{T} p_\theta (\boldsymbol{z})$. We observe that in funnel models samples drawn at lower temperatures are much more structured than those drawn from equivalent lowered temperatures in the NSF model.

Using standard image quality scores we identify that samples drawn from a funnel model have better visual quality than the baseline equivalent, even though the likelihood score is lower. This demonstrates that funnel models perform better at downstream tasks such as image generation. This can further be interpreted as better matching of the encoded data to the prior.
%
\begin{table}
\centering
\caption{Inception score and \textsc{FID} for \textsc{CIFAR-$10$ $5$-bit} for a GLOW based \textsc{NSF} model, a \textsc{F-NSF} model and a ResNet \textsc{VAE}. *Results taken from Ostrovoski et. al \cite{ostrovski2018autoregressive}.}
\begin{tabular}{lcc}
    \toprule \textsc{Model} & \textsc{Inception} $\uparrow$  & \textsc{FID} $\downarrow$ \\
    \cmidrule(r){1-1} \cmidrule(r){2-3} \textsc{DCGAN}* & $6.4$ & $37.1$ \\
    \textsc{WGAN-GP}* & $6.5$ & $36.4$ \\
    \textsc{PixelCNN}* & $4.60$ & $65.93$ \\
    \textsc{PixelQN}* & $5.29$ & $49.46$ \\
    \midrule \textsc{NSF} & $4.15$ & $47.48$ \\
    \textsc{F-NSF} & $\bf{4.95}$ & $\bf{45.52}$ \\
    \textsc{VAE} & $1.00$ & $745$ \\
    \bottomrule
\end{tabular}
\label{table:image_scores}
\end{table}

\textbf{Anomaly Detection.} Out of the box flows often perform poorly on anomaly detection tasks, particularly on images \cite{kirichenko2020normalizing}. While flows do work well in some settings \cite{aarrestad2021dark}, this trend is problematic. By training models on the MNIST dataset of handwritten digits~\cite{mnist} and evaluating the likelihood scores on the fMNIST dataset of fashion products~\cite{xiao2017fashionmnist} we identify that funnel models significantly improve the difference in the likelihood scores, as shown in Table.~\ref{table:ad_scores}. Here we again use the ELBO to provide an out of the box anomaly metric for VAEs. This test demonstrates that funnel models are again more useful for downstream tasks than standard out of the box flows and VAEs. We further remark that a simple \textsc{F-MLP} with spline activations performs significantly better than the \textsc{VAE} model, both of which are constructed from MLPs.
%
%
\begin{table}
\centering
\caption{Bits per dimension scores for models trained on MNIST evaluated on both MNIST and fashion MNIST and the ratio of the likelihoods on both datasets.}
\begin{tabular}{lcccc}
    \toprule \textsc{Model} & \textsc{Latent Size} & \textsc{MNIST} $\downarrow$ & \textsc{f-MNIST} $\uparrow$ & \textsc{Ratio} $\uparrow$ \\
    \cmidrule(r){1-1} \cmidrule(r){2-2} \cmidrule(r){3-5} \textsc{VAE} & $4$ & $7.30$ & $10.40$ & $1.42$ \\
    & $16$ & $6.95$ & $12.66$ & $1.82$ \\
    \textsc{F-NSF} & $4$ & $\boldsymbol{1.21}$ & ${23.42}$ & $\boldsymbol{19.36}$ \\
     & $16$ & $1.28$ & $11.32$ & $8.84$ \\
     \textsc{F-MLP} & $4$ & $2.86$ & $\boldsymbol{30.98}$ & $10.83$ \\
     & $16$ & $2.84$ & $28.59$ & $10.07$ \\
    \midrule \textsc{NSF} & $768$ & $1.10$ & $4.80$ & $4.36$ \\
    \bottomrule
\end{tabular}
\label{table:ad_scores}
\end{table}




\FloatBarrier
\section{Conclusions and discussion}

We have developed the funnel layer, a method that allows flows to be extended to have lower dimensional latent spaces while maintaining exact likelihood estimation and similar or better performance with a similar number of parameters. The reduced latent space of a funnel model is shown to be useful for downstream tasks such as generation and anomaly detection and helps with scalability. 
Crucially the funnel layer allows new transformations to be constructed and could improve the use of exact likelihood methods on tasks that require fast sampling with high dimensional data \cite{krause2021caloflow,Alsing_2018,cosmo2}.

\acksection

The authors would like to acknowledge funding through the SNSF Sinergia grant called Robust Deep Density Models for High-Energy Particle Physics and Solar Flare Analysis (RODEM) with funding number CRSII$5\_193716$. The authors would like to thank Joakim Tutt for useful discussions.

\printbibliography

@misc{durkan2019neural,
      title={Neural Spline Flows}, 
      author={Conor Durkan and Artur Bekasov and Iain Murray and George Papamakarios},
      year={2019},
      eprint={1906.04032},
      archivePrefix={arXiv},
      primaryClass={stat.ML}
}

@misc{kingma2014autoencoding,
      title={Auto-Encoding Variational Bayes}, 
      author={Diederik P Kingma and Max Welling},
      year={2014},
      eprint={1312.6114},
      archivePrefix={arXiv},
      primaryClass={stat.ML}
}

@misc{papamakarios2021normalizing,
      title={Normalizing Flows for Probabilistic Modeling and Inference}, 
      author={George Papamakarios and Eric Nalisnick and Danilo Jimenez Rezende and Shakir Mohamed and Balaji Lakshminarayanan},
      year={2021},
      eprint={1912.02762},
      archivePrefix={arXiv},
      primaryClass={stat.ML}
}

@misc{real_nvp,
      title={Density estimation using Real NVP}, 
      author={Laurent Dinh and Jascha Sohl-Dickstein and Samy Bengio},
      year={2017},
      eprint={1605.08803},
      archivePrefix={arXiv},
      primaryClass={cs.LG}
}

@article{nielsen2020survae,
  title={Survae flows: Surjections to bridge the gap between VAEs and flows},
  author={Nielsen, Didrik and Jaini, Priyank and Hoogeboom, Emiel and Winther, Ole and Welling, Max},
  journal={Advances in Neural Information Processing Systems},
  volume={33},
  year={2020}
}

@misc{uci_datasets,
    author = "Dua, Dheeru and Graff, Casey",
    year = "2017",
    title = "{UCI} Machine Learning Repository",
    url = "http://archive.ics.uci.edu/ml",
    institution = "University of California, Irvine, School of Information and Computer Sciences" 
}

@InProceedings{MartinFTM01,
  author = {D. Martin and C. Fowlkes and D. Tal and J. Malik},
  title = {A Database of Human Segmented Natural Images and its
           Application to Evaluating Segmentation Algorithms and
           Measuring Ecological Statistics},
  booktitle = {Proc. 8th Int'l Conf. Computer Vision},
  year = {2001},
  month = {July},
  volume = {2},
  pages = {416--423}
}

@dataset{papamakarios_george_2018_1161203,
  author       = {Papamakarios, George},
  title        = {Preprocessed datasets for MAF experiments},
  month        = jan,
  year         = 2018,
  publisher    = {Zenodo},
  doi          = {10.5281/zenodo.1161203},
  url          = {https://doi.org/10.5281/zenodo.1161203}
}

@article{papamakarios2017masked,
  title={Masked autoregressive flow for density estimation},
  author={Papamakarios, George and Pavlakou, Theo and Murray, Iain},
  journal={arXiv preprint arXiv:1705.07057},
  year={2017}
}

@article{kingma2018glow,
  title={Glow: Generative flow with invertible 1x1 convolutions},
  author={Kingma, Diederik P and Dhariwal, Prafulla},
  journal={arXiv preprint arXiv:1807.03039},
  year={2018}
}

@misc{huang2020augmented,
      title={Augmented Normalizing Flows: Bridging the Gap Between Generative Flows and Latent Variable Models}, 
      author={Chin-Wei Huang and Laurent Dinh and Aaron Courville},
      year={2020},
      eprint={2002.07101},
      archivePrefix={arXiv},
      primaryClass={cs.LG}
}

@misc{chen2020vflow,
      title={VFlow: More Expressive Generative Flows with Variational Data Augmentation}, 
      author={Jianfei Chen and Cheng Lu and Biqi Chenli and Jun Zhu and Tian Tian},
      year={2020},
      eprint={2002.09741},
      archivePrefix={arXiv},
      primaryClass={stat.ML}
}

@article{tabak2013family,
  title={A family of nonparametric density estimation algorithms},
  author={Tabak, Esteban G and Turner, Cristina V},
  journal={Communications on Pure and Applied Mathematics},
  volume={66},
  number={2},
  pages={145--164},
  year={2013},
  publisher={Wiley Online Library}
}

@misc{nalisnick2019deep,
      title={Do Deep Generative Models Know What They Don't Know?}, 
      author={Eric Nalisnick and Akihiro Matsukawa and Yee Whye Teh and Dilan Gorur and Balaji Lakshminarayanan},
      year={2019},
      eprint={1810.09136},
      archivePrefix={arXiv},
      primaryClass={stat.ML}
}

@misc{rezende2014stochastic,
      title={Stochastic Backpropagation and Approximate Inference in Deep Generative Models}, 
      author={Danilo Jimenez Rezende and Shakir Mohamed and Daan Wierstra},
      year={2014},
      eprint={1401.4082},
      archivePrefix={arXiv},
      primaryClass={stat.ML}
}

@article{kingma2014adam,
  title={Adam: A method for stochastic optimization},
  author={Kingma, Diederik P and Ba, Jimmy},
  journal={arXiv preprint arXiv:1412.6980},
  year={2014}
}

@misc{loshchilov2017sgdr,
      title={SGDR: Stochastic Gradient Descent with Warm Restarts}, 
      author={Ilya Loshchilov and Frank Hutter},
      year={2017},
      eprint={1608.03983},
      archivePrefix={arXiv},
      primaryClass={cs.LG}
}

@misc{ioffe2015batch,
      title={Batch Normalization: Accelerating Deep Network Training by Reducing Internal Covariate Shift}, 
      author={Sergey Ioffe and Christian Szegedy},
      year={2015},
      eprint={1502.03167},
      archivePrefix={arXiv},
      primaryClass={cs.LG}
}

@article{hahnloser2000digital,
  title={Digital selection and analogue amplification coexist in a cortex-inspired silicon circuit},
  author={Hahnloser, Richard HR and Sarpeshkar, Rahul and Mahowald, Misha A and Douglas, Rodney J and Seung, H Sebastian},
  journal={Nature},
  volume={405},
  number={6789},
  pages={947--951},
  year={2000},
  publisher={Nature Publishing Group}
}

@misc{hoogeboom2019emerging,
      title={Emerging Convolutions for Generative Normalizing Flows}, 
      author={Emiel Hoogeboom and Rianne van den Berg and Max Welling},
      year={2019},
      eprint={1901.11137},
      archivePrefix={arXiv},
      primaryClass={cs.LG}
}

@misc{oliva2018transformation,
      title={Transformation Autoregressive Networks}, 
      author={Junier B. Oliva and Avinava Dubey and Manzil Zaheer and Barnabás Póczos and Ruslan Salakhutdinov and Eric P. Xing and Jeff Schneider},
      year={2018},
      eprint={1801.09819},
      archivePrefix={arXiv},
      primaryClass={stat.ML}
}

@misc{krause2021caloflow,
      title={CaloFlow: Fast and Accurate Generation of Calorimeter Showers with Normalizing Flows}, 
      author={Claudius Krause and David Shih},
      year={2021},
      eprint={2106.05285},
      archivePrefix={arXiv},
      primaryClass={physics.ins-det}
}

@article{Alsing_2018,
   title={Massive optimal data compression and density estimation for scalable, likelihood-free inference in cosmology},
   volume={477},
   ISSN={1365-2966},
   url={http://dx.doi.org/10.1093/mnras/sty819},
   DOI={10.1093/mnras/sty819},
   number={3},
   journal={Monthly Notices of the Royal Astronomical Society},
   publisher={Oxford University Press (OUP)},
   author={Alsing, Justin and Wandelt, Benjamin and Feeney, Stephen},
   year={2018},
   month={Mar},
   pages={2874–2885}
}

@article{cosmo2,
    author = {Rouhiainen, Adam and Giri, Utkarsh and M\"unchmeyer, Moritz},
    title = "{Normalizing flows for random fields in cosmology}",
    eprint = "2105.12024",
    archivePrefix = "arXiv",
    primaryClass = "astro-ph.CO",
    month = "5",
    year = "2021"
}

@inproceedings{ostrovski2018autoregressive,
  title={Autoregressive quantile networks for generative modeling},
  author={Ostrovski, Georg and Dabney, Will and Munos, R{\'e}mi},
  booktitle={International Conference on Machine Learning},
  pages={3936--3945},
  year={2018},
  organization={PMLR}
}

@misc{kirichenko2020normalizing,
      title={Why Normalizing Flows Fail to Detect Out-of-Distribution Data}, 
      author={Polina Kirichenko and Pavel Izmailov and Andrew Gordon Wilson},
      year={2020},
      eprint={2006.08545},
      archivePrefix={arXiv},
      primaryClass={stat.ML}
}

@misc{aarrestad2021dark,
      title={The Dark Machines Anomaly Score Challenge: Benchmark Data and Model Independent Event Classification for the Large Hadron Collider}, 
      author={T. Aarrestad and M. van Beekveld and M. Bona and A. Boveia and S. Caron and J. Davies and A. De Simone and C. Doglioni and J. M. Duarte and A. Farbin and H. Gupta and L. Hendriks and L. Heinrich and J. Howarth and P. Jawahar and A. Jueid and J. Lastow and A. Leinweber and J. Mamuzic and E. Merényi and A. Morandini and P. Moskvitina and C. Nellist and J. Ngadiuba and B. Ostdiek and M. Pierini and B. Ravina and R. Ruiz de Austri and S. Sekmen and M. Touranakou and M. Vaškevičiūte and R. Vilalta and J. R. Vlimant and R. Verheyen and M. White and E. Wulff and E. Wallin and K. A. Wozniak and Z. Zhang},
      year={2021},
      eprint={2105.14027},
      archivePrefix={arXiv},
      primaryClass={hep-ph}
}

@misc{oord2016pixel,
      title={Pixel Recurrent Neural Networks}, 
      author={Aaron van den Oord and Nal Kalchbrenner and Koray Kavukcuoglu},
      year={2016},
      eprint={1601.06759},
      archivePrefix={arXiv},
      primaryClass={cs.CV}
}

@misc{he2015deep,
      title={Deep Residual Learning for Image Recognition}, 
      author={Kaiming He and Xiangyu Zhang and Shaoqing Ren and Jian Sun},
      year={2015},
      eprint={1512.03385},
      archivePrefix={arXiv},
      primaryClass={cs.CV}
}

@article{mnist,
  author={Lecun, Y. and Bottou, L. and Bengio, Y. and Haffner, P.},
  journal={Proceedings of the IEEE}, 
  title={Gradient-based learning applied to document recognition}, 
  year={1998},
  volume={86},
  number={11},
  pages={2278-2324},
  doi={10.1109/5.726791}}

@misc{xiao2017fashionmnist,
      title={Fashion-MNIST: a Novel Image Dataset for Benchmarking Machine Learning Algorithms}, 
      author={Han Xiao and Kashif Rasul and Roland Vollgraf},
      year={2017},
      eprint={1708.07747},
      archivePrefix={arXiv},
      primaryClass={cs.LG}
}

@techreport{Krizhevsky09learningmultiple,
    author = {Alex Krizhevsky},
    title = {Learning multiple layers of features from tiny images},
    institution = {},
    year = {2009}
}

@misc{russakovsky2015imagenet,
      title={ImageNet Large Scale Visual Recognition Challenge}, 
      author={Olga Russakovsky and Jia Deng and Hao Su and Jonathan Krause and Sanjeev Satheesh and Sean Ma and Zhiheng Huang and Andrej Karpathy and Aditya Khosla and Michael Bernstein and Alexander C. Berg and Li Fei-Fei},
      year={2015},
      eprint={1409.0575},
      archivePrefix={arXiv},
      primaryClass={cs.CV}
}

@misc{ziegler2019latent,
      title={Latent Normalizing Flows for Discrete Sequences}, 
      author={Zachary M. Ziegler and Alexander M. Rush},
      year={2019},
      eprint={1901.10548},
      archivePrefix={arXiv},
      primaryClass={stat.ML}
}

\clearpage
\appendix

\section{Appendix}

\subsection{Definitions and notation}
\label{app:defs_and_notation}

For the bound looseness term to be zero the right inverse condition must be satisfied. The right inverse condition can be stated as $f(\boldsymbol{x}) = \boldsymbol{z}$ for any $\boldsymbol{x} \sim p(\boldsymbol{x}|\boldsymbol{z})$ and $\boldsymbol{z} \in \mathbb{Z}$, where $\mathbb{Z}$ is the support of $q(\boldsymbol{z}) = \mathbb{E}_{p_D(\boldsymbol{x})}[q(\boldsymbol{z}|\boldsymbol{x})]$. Funnel models satisfy this condition because

\begin{align*}
    p(\xPlus|\boldsymbol{z}) = \delta (\xPlus - F(\boldsymbol{z})) &\implies F(\boldsymbol{z}) = \xPlus \sim p(\xPlus|\boldsymbol{z})\\
    &\implies F^{-1}(\xPlus) = \boldsymbol{z}, ~\forall \xPlus \sim p(\xPlus|\boldsymbol{z}).\\
\end{align*}

Note also that a diffeomorphism $F$ parameterized by a function $\theta(\xMinus, \xPlus)$ is defined such that its inverse can be found given $\boldsymbol{z}$ using $\theta(\xMinus, \boldsymbol{z})$. This is done  by using coupling or autoregressive transformations on $\xPlus$.

\subsection{Generative funnel layer}
\label{app:derivation_of_loss}

We can also develop a generative funnel surjection. The likelihood contribution for a generative surjection is given by
\begin{equation*}
    \mathcal{V}(\boldsymbol{x}, \boldsymbol{z}) = \lim_{p(\boldsymbol{x} | \boldsymbol{z}) \rightarrow \delta (\boldsymbol{x} - f^{-1}(\boldsymbol{z}))} \mathbb{E}_{q(\boldsymbol{z}|\boldsymbol{x})} \left[ \log \frac{ p(\boldsymbol{x} | \boldsymbol{z}) }{ q(\boldsymbol{z}|\boldsymbol{x}) } \right].
    \label{eq:likelihood_contribution_gener}
\end{equation*}
For a generative funnel transformation we split $\boldsymbol{z} = \{ \boldsymbol{z}_{1:n}, \boldsymbol{z}_{n+1:m} \}$ and use a diffeomorphism $G$ such that $\boldsymbol{z}_{1:n} = G^{-1}_{\theta(\boldsymbol{x}, \boldsymbol{z}_{n+1:m})}(\boldsymbol{x})$ with $\boldsymbol{z}_{n+1:m} \sim p(\boldsymbol{z}_{n+1:m} | \boldsymbol{x})$. Then following the same steps as for the inference surjection we can use
\begin{align*}
    p(\boldsymbol{x} | \boldsymbol{z}) &= \delta (\boldsymbol{x} - G_{\theta(\boldsymbol{z}_{1:n}, \boldsymbol{z}_{n+1:m})}(\boldsymbol{z}_{1:n}))\\
     &= \delta (\boldsymbol{z}_{1:n} - G^{-1}_{\theta(\boldsymbol{z}_{1:n}, \boldsymbol{z}_{n+1:m})}(\boldsymbol{x})) \left| \det (J_\theta (\boldsymbol{z}) ) \right|,
\end{align*}
to write
\begin{align}
    \begin{split}
    \mathcal{V}(\boldsymbol{x}) 
    &= \int \dif \boldsymbol{z} q(\boldsymbol{z} | \boldsymbol{x}) \log \left( \frac{\delta (\boldsymbol{z}_{1:n} - G_{\theta(\boldsymbol{z}_{1:n}, \boldsymbol{z}_{n:m})}(\boldsymbol{x})) \left| \det (J_\theta (\boldsymbol{z}) ) \right|}{ q(\boldsymbol{z} | \boldsymbol{x}) } \right)\\
    &= \int \dif \boldsymbol{z}_{n+1:m} q(\boldsymbol{z}_{n+1:m} | \boldsymbol{x}) \log \left( \frac{\left| \det (J_\theta (\boldsymbol{z}) ) \right|}{q(\boldsymbol{z}_{n+1:m} | \boldsymbol{x})} \right),\\
    \end{split}
    \label{eq:gen_sur}
\end{align}
with
\begin{equation*}
    J_\theta (\boldsymbol{z}) = \left. \frac{\partial G^{-1}_{\theta(\boldsymbol{z}_{1:n}, \boldsymbol{z}_{n+1:m})}(\boldsymbol{x})}{\boldsymbol{x}} \right|_{\boldsymbol{x} = G_{\theta(\boldsymbol{z}_{1:n}, \boldsymbol{z}_{n+1:m})}(\boldsymbol{z}_{1:n})}.
\end{equation*}

\subsection{Relationship of funnel to slice surjection}
\label{app:relation_to_slice}

The funnel layer is related to the slice surjection presented in Ref.~\cite{nielsen2020survae}. There are several key differences between the slice surjection and the funnel layer, but before discussing them we detail how the slice surjection can be related to the funnel when composing with invertible neural networks (INNs).

Define a coupling INN $F_{\theta_s}$ that acts on $\boldsymbol{x}$
\begin{equation}
    F_{\theta_s}(\boldsymbol{x}) = (\xMinus, F_{\theta_s(\xMinus)}(\xPlus)),
    \label{eq:specific_inn}
\end{equation}
and a likelihood contribution of
\begin{equation}
    \mathcal{V}(\boldsymbol{x}) =  \log ( | \mathrm{det} J_{F_{\theta_s}} (\xPlus) | ).
\end{equation}
Combining this with a slice surjection $S$ that acts as
\begin{equation}
    S(F_{\theta_s}(\boldsymbol{x})) = F_{\theta_s(\xMinus)}(\xPlus),
\end{equation}
the total likelihood contribution will be
\begin{equation}
    \mathcal{V}(\boldsymbol{x}) = \log \left(  p \left(\xMinus \big| F_{\theta_s(\xMinus)}(\xPlus) \right) \right) + \log ( | \mathrm{det} J_{F_{\theta_s}} (\xPlus) | ).
\end{equation}

This is a special case of the likelihood contribution in Eq.~\eqref{eq:new_loss}, and it requires selecting a particular invertible neural network in Eq.~\eqref{eq:specific_inn}. The funnel layer provides a unified perspective on this combination and allows for more flexibility in terms of how the dependence on $\xMinus$ is propagated as well as the choice of the function $\fForward$ that is used in the transformation.

The slice layer is restricted to being composed with INNs for the function $F_{\theta_s}$. This is not the case for the funnel method, and this freedom allows convolutional kernels to be used with maximum likelihood estimation within the funnel framework as we show in Appendix~\ref{app:conv_kernels}. In general the funnel method requires that $\fForward$ is invertible for $\xPlus$ given $\xMinus$ and has a tractable determinant, this includes functions that are not typically considered as INNs.

When the INN in Eq.~\eqref{eq:specific_inn} is parameterized in terms of a two layer neural network $\theta = \theta_1 \circ \theta_0$ that acts on $(\xMinus, \xPlus)$ the dependence on $\xMinus$ is propagated as
\begin{equation}
    \theta(\xMinus, \xPlus) = \theta_1 (\theta_0 (\xMinus, \xPlus) ),
\end{equation}
which is the standard way of building a coupling INN. In comparison, when using the funnel method the above would be possible, but it is also natural to consider building the dependence as
\begin{equation}
    \theta(\xMinus, \xPlus) = \theta_1 (\xMinus, \theta_0 (\xMinus, \xPlus) ),
    \label{eq:funnel_dep}
\end{equation}
where the resulting transformation has an increased dependence on $\xMinus$. In comparison the INN in Eq.~\eqref{eq:specific_inn} $\xMinus$ would normally be transformed in the next layer of the model, and so it would be unnatural to propagate the dependence on $\xMinus$ in the way shown in Eq.~\ref{eq:funnel_dep}.

Another benefit of the funnel method is clarity of thinking about the different layers. By building the generic composition directly the structure of the model can be viewed holistically. 

The slice surjection can be viewed as a particular case of the funnel where the forward transformation $F$ is the identity. Further, any dependence that can be built into a slice surjection through composition with INNs can also be built into a funnel layer using the same composition.

The inference version of the slice surjection was related to the multi-scale architecture from Ref.~\cite{real_nvp}. Here we demonstrate that its generalisation, the funnel, can be used to build dimension reducing transformations which are effective on a wide variety of datasets.

\subsection{VAE ELBO}
\label{app:vae_elbo}
The evidence lower bound (ELBO) of a variational autoencoder (VAE) is given by
\begin{equation}
    \log p(\boldsymbol{x}) = \mathbb{E}_{q(\boldsymbol{z} | \boldsymbol{x})} [\log p(\boldsymbol{x} | \boldsymbol{z})] - \mathbb{D}_{\mathrm{KL}}[q(\boldsymbol{z} | \boldsymbol{x}) || p(\boldsymbol{z})].
\end{equation}

In this work we calculate the KL divergence term with the reparameterisation trick as is standard, and the expectation value is estimated with a single sample and we assume a gaussian distribution for the reconstruction using a neural network $g_\theta(\boldsymbol{z})=[\mu(\boldsymbol{z}),\sigma(\boldsymbol{z})]$ such that
\begin{equation}
    p(\boldsymbol{x} | \boldsymbol{z}) = \frac{1}{\sigma(\boldsymbol{z}) \sqrt{2\pi}} e^{-\frac{1}{2} \left( \frac{\boldsymbol{x} -\mu(\boldsymbol{z})}{\sigma(\boldsymbol{z})} \right)^2}.
\end{equation}

\subsection{Plane data distribution}
\label{app:plane_data}

\begin{figure}
    \centering
    \includegraphics[scale=0.2]{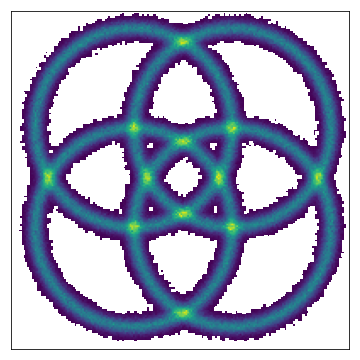}
    \caption{The data distribution (Four circles) used to train the model in Fig.~\ref{fig:quantitative}.}
    \label{fig:four_circles}
\end{figure}

The datatset that is modelled in Fig.~\ref{fig:quantitative} is called four circles and the data distribution is shown in Fig.~\ref{fig:four_circles}.

On this dataset a neural spline flow (NSF) \cite{durkan2019neural} with the last layer replaced by a funnel was used. The NSF was formed from two autoregressive rational quadratic layers with $50$ bins in each layer, a tail bound of $4$ and a standard normal as the base distribution. The parameters of the splines are learned with two residual blocks composed of two multilayer perceptrons (MLP) with $256$ hidden units. The inverse density $p_\phi(x_1 | \boldsymbol{z})$ is composed of an NSF with the same structure but with $64$ nodes in the MLP and ten bins per layer.

\subsection{UCI experimental details}
\label{app:uci_details}

All parameters are chosen to replicate the experiments from Ref.~\cite{durkan2019neural} and funnel models are built such that they have less parameters than their non-funnel baselines. No model selection has been performed for the funnel approaches, except tuning to match the numbers of parameters used. The norm of the gradients is clipped to the range $[-5, 5]$. Hyperparameter settings are shown for coupling flows in Table~\ref{tab:tablular_data_setup}. The dimensionality and number of training data points in each dataset is included for reference. When using a flow to parameterize the density $p(\xMinus | \boldsymbol{z})$ the same type of flow as is used in the modelling is used, with half the number of flow steps and one third as many hidden features in the residual blocks. For BSDS$300$ a Gaussian ansatz was used for this density, with an MLP composed of three layer with $128$ nodes and ReLU activations~\cite{hahnloser2000digital} to predict the mean and diagonal covariance matrix of the Gaussian.

\begin{table*}[h]
    \centering
    \caption{The hyperparameters used for uci datasets and BSDS$300$ for both flows and MLPs.}
    \resizebox{\textwidth}{!}{
    \begin{tabular}{lccccc}
        \toprule & \textsc{POWER} & \textsc{GAS} & \textsc{HEPMASS} & \textsc{MINBOONE} & \textsc{BSDS$300$}\\
        \cmidrule(r){1-6} \textsc{Dimension} & $6$ & $8$ & $21$ & $43$ & $63$ \\ 
        \textsc{Data Points} & $1{,}615{,}917$ & $852{,}174$ & $315{,}123$ & $29{,}556$ & $1{,}000{,}000$ \\ 
        \cmidrule(r){1-6} \textsc{Batch Size} & $512$ & $512$ & $256$ & $128$ & $512$ \\
        \textsc{Training Steps} & $400{,}000$ & $400{,}000$ & $400{,}000$ & $200{,}000$ & $400{,}000$ \\
        \textsc{Learning Rate} & $0.0005$ & $0.0005$ & $0.0005$ & $0.0003$ & $0.0005$ \\
        \textsc{Flow Steps} & $10$ & $10$ & $20$ & $10$ & $10$ \\
        \textsc{Residual Blocks} & $2$ & $2$ & $1$ & $1$ & $1$ \\
        \textsc{Hidden Features} & $256$ & $256$ & $128$ & $32$ & $128$ \\
        \textsc{Bins} & $8$ & $8$ & $8$ & $4$ & $8$ \\
        \textsc{Dropout} & $0.0$ & $0.1$ & $0.2$ & $0.2$ & $0.2$ \\
        \textsc{Dim. Reduction} & $2$ & $2$ & $5$ & $10$ & $15$ \\
        \textsc{Funnel Level} & $2$ & $2$ & $0$ & $4$ & $4$ \\
        \cmidrule(r){1-6} \textsc{VAE} & & & & & \\
        \cmidrule(r){1-1} \textsc{MLP Width} & $256$ & $64$ & $256$ & $512$ & $512$ \\
        \textsc{MLP Depth} & $2$ & $8$ & $2$ & $2$  & $2$ \\
        \textsc{Learning Rate} & $$0.003$$ & $0.005$ & $0.005$ & $0.001$ & $0.003$ \\
        \bottomrule
    \end{tabular}
    }
    \label{tab:tablular_data_setup}
\end{table*}

For the VAEs used in these experiments a hyper parameter scan was performed to find the best model, with the number of parameters varying across the scan but allowing for both more and less parameters than the flow models. In all cases the \textsc{Dropout}$=0$ and the \textsc{Batch Size} was the same as that used for flows. 
Both batch normalization and layer normalization were included in the scan but never resulted in the best performance. The parameters in Table.~\ref{table:grid_search} were searched to find the best VAE for each dataset.

\begin{table}
\caption{The parameters searched to find the best VAE for each of the tabular datasets.}
\centering
    \begin{tabular}{ll}
    \toprule
     \textsc{MLP Width} & $64, 128, 256, 512$ \\
     \textsc{MLP Depth} & $2, 3, 4, 6, 8$ \\
     \textsc{Dropout} & $0.0, 0.2, 0.5$ \\
     \textsc{Batch Size} & $128, 256, 512, 1024$ \\
     \textsc{Learning Rate} & $0.0003, 0.0001, 0.00008, 0.00005$ \\
     \bottomrule
     \end{tabular}
     \label{table:grid_search}
\end{table}

Each layer of the \textsc{F-MLP} model used a fixed random permutation followed by an \textsc{F-MLP} layer where an invertible matrix parameterized by the LU decomposition was used for $\boldsymbol{R}$. For dimension preserving layers just the matrix $\boldsymbol{R}$ was used. A rational quadratic spline with $10$ bins and a tail bound of $2$ was used as an activation for each layer. Each experiment used two dimension preserving layers, followed by one layer to map to a vector with half the dimension of the input and the latent space dimension, followed by a dimension preserving layer, a layer to map to the latent space, and then three dimension preserving layers. All other parameters were held the same as those used for the flows in Table.~\ref{tab:tablular_data_setup} and no hyper parameter scan was performed.


\subsection{Standard layers}
\label{app:funneling_standard_layers}

In this section we detail how both standard convolutional kernels and MLPs can be converted into funnel models.

\subsubsection{Convolutional kernel funnels}
\label{app:conv_kernels}

In this section we outline the method for turning convolutional kernels into a funnel layer. We will recover the simple one dimensional example and then extend to more complex data types.

Given a one dimensional vector $\boldsymbol{x} \in \mathbb{R}^{n}$ the forward transformation of the funnel can be factorised to form a $2\times 1$ convolutional kernel ($[a, b]$) with stride two as follows
\begin{equation}
    q (\boldsymbol{z} | \boldsymbol{x}) = \prod_{i=1}^{n/2} q (z_i | x_{2i - 1}, x_{2i})
\end{equation}
with,
\begin{align*}
    z_i &= F^{-1}_{\theta(x_{2i - 1}, x_{2i})}(x_{2i - 1}) \\
    &= a x_{2i - 1} + b x_{2i}
\end{align*}
where $[a, b]$, are the parameters of the convolutional kernel. The Jacobian of each individual kernel is given by
\begin{align*}
    J_i &= \left[\frac{\dif F^{-1}_{\theta(x_{2i - 1}, x_{2i})}(x_{2i - 1})}{ \dif x_{2i - 1}}\right]= [a],\\
\end{align*}
and the total Jacobian will be given by $\frac{n}{2}a$ such that the total likelihood contribution is given by
\begin{align*}
    \mathcal{V}(\boldsymbol{x}) = \log \left(  p_\phi \left( \{ x_{2i} \}_{i=1}^{n/2} \big| \boldsymbol{z} \right) \right) + \log\left(\frac{n}{2}| a |\right).
\end{align*}
Given $x_{2i}' \sim p_\phi \left( \{ x_{2i} \}_{i=1}^{n/2} | \boldsymbol{z} \right)$, the kernel $[a, b]$ and $\boldsymbol{z}$ we can find $x_{2i - 1}$
\begin{equation}
    x_{2i - 1} = J_i^{-1} (z_i - b x_{2i}'),
\end{equation}
for all $i$ such that the right inverse condition is satisfied. Here the different elements of the kernel play different roles in the transformation with $a$ appearing directly in the loss and $b$ playing an implicit role in $p_\phi$, we also chose $x_{2i - 1}$ arbitrarily to play the role of $\xPlus$. As we have chosen a stride of two, this means different features in the input vector will also play different roles. The issue of certain features playing a special role is resolved if we take a stride of one, as then the forward transformation is given by
\begin{equation*}
    q(z_i | x_{i}, x_{i + 1}) = \delta ( z_i - a x_{i} - b x_{i+1} ),
\end{equation*}
for all $i\in[1, n-1]$, and all features are multiplied by the special element of the kernel ($a$). But, in this setting we have the autoregressive property and so 
\begin{equation*}
    p_\phi \left( \boldsymbol{x} | \boldsymbol{z} \right) = p_\phi \left( x_{n} \big| \boldsymbol{z} \right),
\end{equation*}
because if we have access to $x_n$ (or a sample from $p_\phi \left( x_{n} \big| \boldsymbol{z} \right)$) then $\boldsymbol{x}$ can be unambiguously recovered using the recursion relation
\begin{equation*}
    x_{n-i} = J_i^{-1} (z_{n - i} - b x_{n - i + 1}'),
\end{equation*}
for all $i\in[1, n-1]$. In settings with low dimensional data this is a very useful property, but for images it makes the sampling process very slow. To speed up sampling the likelihood of the model can be bounded from below by violating the right inverse condition, and estimating $p_\phi \left( \boldsymbol{x} \big| \boldsymbol{z} \right)$ directly. To match the appearance of $a$ in the likelihood contribution it is possibly better to have a generative process $p_\phi$ that factorises as
\begin{align*}
    p_\phi \left( \boldsymbol{x} \big| \boldsymbol{z} \right) &= p_\phi(x_1 | a x_1 + b x_2) \prod_{i=2}^n p_\phi \left( x_i \big| a x_{i - 1} + b x_{i} \right),
\end{align*}
where the conditioning in all densities except the first enhances the role played by $b$. This is the form of the generative process that we use in all studies, were the same model $p_\phi$ is used in each inverse block. A better approach would be to remove the ambiguity associated with the choice of kernel element to appear in the loss.

The above approach can be extended to three channel images. Given a kernel defined by $a_{i, j}^{k,l}$, where $a \in \mathbb{R}$ and $i, j$ are the indices of each matrix in the kernel, $k$ corresponds to the number of output channels and $l$ to the number of input channels. Fixing $k = l$ and $i = j = 2$ and choosing (again arbitrarily) to take the bottom right component of each matrix to be $\xPlus$ the Jacobian is given by
\begin{equation}
    J = \begin{bmatrix}
    a_{2, 2}^{1, 1} & a_{2, 2}^{2, 1} & a_{2, 2}^{3, 1} \\
    a_{2, 2}^{1, 2} & a_{2, 2}^{2, 2} & a_{2, 2}^{3, 2} \\
    a_{2, 2}^{1, 3} & a_{2, 2}^{2, 3} & a_{2, 2}^{3, 3} \\
    \end{bmatrix}.
\end{equation}
The determinant of the Jacobian has to be calculated once per batch, and so allows for efficient training. The inverse Jacobian has to be calculated once for every trained model, allowing for efficient sampling. 

Samples can be generated efficiently using the following steps. Defining the kernel defined by $a_{i, j}^{k,l}$ to act as $F^{-1}(\xMinus, \xPlus)$ then for each tile $j$ in the forward convolution we can find the coordinates $\xPlus^j$ by sampling $\xMinus^j$ from $p_\phi(\xMinus | z_j)$ and using
\begin{equation}
    \boldsymbol{x}_{m:n}^j = J^{-1}(z_j - F^{-1}(\xMinus^j ,\boldsymbol{0}_{\mathbb{R}^3}) ).
\end{equation}
to ensure the right inverse condition will be satisfied in the case where the stride of the kernel is equivalent to its width. This procedure can be parallelised across all tiles of a convolution in standard machine learning libraries, leading to an efficient sampling process.

This generalizes to all $\{i, j\}$ but the requirement that $k = l$ is fixed to ensure the matrix $J$ is square. The same considerations apply when using a stride that is less than the width of the kernel.

Other approaches with funnel models can be developed, such as using coupling transformations and dropping some of the untransformed channels after the transformation has been applied. In practise this works but requires more parameters to function and so was not included in the comparisons in Table.~\ref{tab:image_data_results}. This approach could be used, for example, to transform grey scale to color images.

\subsubsection{MLP funnels}
\label{app:MLPfunnels}

Dimension increasing funnels cannot be developed within this framework as the inference surjection is needed, but this requires a stochastic $q(\boldsymbol{z} | \boldsymbol{x})$ which cannot be done with standard MLPs.

\subsection{Image experimental details}
\label{app:image_exp_dets}

For the generative modelling of images experiments we use a rational quadratic spline based architecture \cite{durkan2019neural} that is based on a Glow-like model \cite{kingma2018glow}. Our approach exactly replicates all experimental details from Ref.~\cite{durkan2019neural}, where the multi-scale architecture that is used here can be viewed as a simple slice surjection \cite{nielsen2020survae}. In the following we replicate the information provided in Ref.~\cite{durkan2019neural}


The architecture consist of multiple steps for each level in a multiscale architecture \cite{real_nvp}, where each step consists of an actnorm layer, an invertible $1 \times 1$ convolution and a coupling transform. For our RQ-NSF (C) model, rational quadratic coupling transforms parameterized by residual convolutional networks are used, and a $1 \times 1$ convolution at the end of each level of transform. For CIFAR-$10$ experiments dimensions are not factored out at the end of each level, but the squeezing operation is used to trade spatial resolution for depth. All experiments used $3$ residual blocks and batch normalization~\cite{ioffe2015batch} in the residual networks which parameterize the coupling transforms. We use $7$ steps per level for all experiments, resulting in a total of $21$ coupling transforms for CIFAR-$10$, and $28$ coupling transform for ImageNet$64$.

We use the Adam optimizer~\cite{kingma2014adam} with default $\beta_1$ and $\beta_2$ values. An initial learning rate of $0.0005$ is annealed to $0$ following a cosine schedule~\cite{loshchilov2017sgdr}. We train for $100{,}000$ steps for $5$-bit experiments, and for $200{,}000$ steps for $8$-bit experiments. Final values for all hyperparameters are reported in Table~\ref{tab:image_data_setup}. We use a single Nvidia RTX $3090$ GPU card per image experiment. Training for $200,000$ steps takes about $4$ days on this card. All funnel experiments are run for $12$ hours.

As we are interested in dimension reducing transformations, we only used convolutions where the width of the kernel and the stride are equivalent. In all cases we use a convolutional kernel of size $2$ with $3$ channels and no padding. We apply an actnorm layer, an invertible $1 \times 1$ convolution and a coupling transformation (the first block of the NSF architecture) followed by the convolutional kernel and then the rest of the NSF architecture. The Gaussian ansatz, used to parameterize the log likelihood of the tiles in the kernel, is formed from a three layer MLP with $128$ nodes in each layer and ReLU activations. In this case the ansatz is trained to invert the tile of each kernel individually, given all neighbours of the corresponding element $z_i$. This results in a tile of $z_i$ elements of size $3$ with $3$ channels giving $27$ inputs to infer the $(2 \times 2 - 1)\times 3 = 9$ elements of $\xMinus$ in each tile in the input image.

\begin{table*}[h]
    \centering
    \caption{The hyperparameters used for generative modelling of images.}
    \resizebox{\textwidth}{!}{ 
    \begin{tabular}{lcccccc}
        \toprule \textsc{Dataset} & \textsc{Bits} & \textsc{Batch Size} & \textsc{Levels} & \textsc{Hidden Channels} & \textsc{Bins} & \textsc{Dropout}\\
        \cmidrule(r){1-7} \textsc{CIFAR}-$10$ & $5$ & $512$ & $3$ & $64$ & $2$ & $0.2$ \\ 
         & $8$ & $512$ & $3$ & $96$ & $4$ & $0.2$ \\ 
         \cmidrule(r){2-7} \textsc{ImageNet}$64$ & $5$ & $256$ & $4$ & $96$ & $8$ & $0.1$ \\ 
         & $8$ & $256$ & $4$ & $96$ & $8$ & $0.0$ \\ 
        \bottomrule
    \end{tabular}
    }
    \label{tab:image_data_setup}
\end{table*}

For the VAEs in these experiments a ResNet$18$ encoder \cite{he2015deep} with a single linear layer to map from the output of the ResNet$18$ model to the latent space. The same convolutions are repeated for the decoder with convolutional kernels for the upsampling.

\subsection{Anomaly detection}
For the \textsc{NSF} model we use rational quadratic coupling transforms parameterized by residual convolutional networks, and a $1\times1$ convolution at the end of each level of transform. All experiments use $3$ residual blocks and batch normalization \cite{ioffe2015batch} and $7$ steps per level. Squeezing is applied between different layers to exchange spatial resolution for depth.

For the \textsc{F-NSF} we used a convolution of size $(2, 2)$ and stride two in each layer, with zero padding after the first convolution to ensure the dimension of the image remains even. To compress to to $16$ dimensions three layers were used, and to compress to $4$ dimensions four layers were used. The same layers were used as in the case of the \textsc{NSF} but no squeezing was applied.

The \textsc{F-MLP} model used a fixed random permutation and an invertible matrix parameterized by the LU decomposition was used for $\boldsymbol{R}$. A fixed per dimension rational quadratic spline with $5$ bins and a tial bound of $4$ \cite{durkan2019neural} was used as an activation and the network compressed the input by first flattening the image and then applying five \textsc{F-MLP} layers as $768\rightarrow512\rightarrow256\rightarrow128\rightarrow64\rightarrow\mathrm{LS}$ where the latent space size ($\mathrm{LS}$) is either $4$ or $16$.

The \textsc{VAE} model was parameterized by linear layers with \textsc{ReLU} activations \cite{hahnloser2000digital}, the encoder and decoder had the same structure and the network compressed the image by first flattening and then applying linear layers as $768\rightarrow512\rightarrow512\rightarrow512\rightarrow\mathrm{LS}$ where $\mathrm{LS}$ is either $4$ or $16$. 

\begin{table}
\centering
\caption{The number of parameters used in each of the models in Table.~\ref{table:ad_scores}.}
\begin{tabular}{lcc}
    \toprule \textsc{Model} & \textsc{Latent Size} & \textsc{Param} \\
    \cmidrule(r){1-1} \cmidrule(r){2-3} \textsc{VAE} & $4$ & $1{,}861{,}401$ \\
    & $16$ & $1{,}879{,}857$  \\
    \textsc{F-NSF} & $4$ & $7{,}286{,}164$  \\
     & $16$ & $5{,}464{,}623$ \\
     \textsc{F-MLP} & $4$ & $1{,}890{,}324$  \\
     & $16$ & $1{,}888{,}020$ \\
    \midrule \textsc{NSF} & $768$ & $4{,}724{,}430$  \\
    \bottomrule
\end{tabular}
\label{table:ad_param}
\end{table}

\end{document}